%% file: main.tex
\documentclass{article}

\usepackage{arxiv}

\usepackage[utf8]{inputenc} %
\usepackage[T1]{fontenc}    %
\usepackage{hyperref}       %
\usepackage{url}            %
\usepackage{booktabs}       %
\usepackage{amsfonts}       %
\usepackage{nicefrac}       %
\usepackage{microtype}      %
\usepackage{graphicx}
\usepackage{natbib}
\usepackage{doi}
\usepackage{amssymb}
\usepackage{amsmath}
\usepackage{subfigure}
\usepackage{makeidx}
\usepackage{multicol}

\usepackage{etoolbox} %

\newcommand{\pasthorizon}{10\,}
\title{High-frequency pricing at scale for e-commerce}

\author{{\hspace{1mm}Stefan Birr}\\
        Zalando SE\thanks{Contact: stefan.birr@zalando.de}
	\And
	\hspace{1mm}Tobias Huelden\\
        Zalando SE
        \And
        {\hspace{1mm}Mones Raslan}\\
        Zalando SE
        \And
        {\hspace{1mm}Adele Gouttes}\\
        Zalando SE
        \And
        {\hspace{1mm}Andreas Schmitt}\\
        Zalando SE
        \And
        {\hspace{1mm}Mateusz Koren}\\
        Zalando SE
        \And
        {\hspace{1mm}Johannes Stephan}\\
        Zalando SE
        \And
        {\hspace{1mm}Robert Streek}\\
        Zalando SE
        \And
        {\hspace{1mm}Manuel Kunz}\\
        Zalando SE
        \And
        {\hspace{1mm}Tim Januschowski}\\
        Databricks\thanks{Work done while working at Zalando.}
}

\renewcommand{\shorttitle}{High-frequency pricing at scale for e-commerce}

\newcommand{\tch}[1]{\TableColHeadFont 1\llstrut\hfill}

\newcommand\llstrut{\rule[-6pt]{0pt}{14pt}}

\hypersetup{
pdftitle={High-frequency pricing at scale for e-commerce},
pdfkeywords={demand forecasting, pricing, sales events, online experimentation},
}

\begin{document}
\maketitle

\begin{abstract}
	This paper presents the design, development, and implementation of a specialized forecast-then-optimize algorithmic pricing tool for sales campaigns in fashion e-commerce. Sales events present unique challenges for pricing including volatile demand patterns, rapid pricing decisions, and the need to balance short-term revenue with long-term profitability. We describe our approach combining daily-resolution demand forecasting using gradient-boosted trees with a multi-objective optimization framework that maximizes both long-term profit and net merchandise value for more than 5 million articles.
Our solution addresses key limitations of existing weekly-granularity systems by implementing a forecast-then-optimize architecture that reduces pricing decision time from hours to minutes.
We validate our approach through 23 A/B tests across 12 markets during 2023-2024 sales campaigns at Zalando, one of Europe's leading online fashion retailers. Experimental results demonstrate that the new pricing system achieves approximately 6\% higher profit while maintaining equivalent performance on sales and revenue compared to the previous manual-algorithmic hybrid approach.
Based on these results, the algorithm was successfully deployed to production and now handles the majority of algorithmic pricing decisions for sales campaigns at the company.
\end{abstract}


\section{Introduction}
Zalando is one of Europe's leading e-commerce companies, specializing in fashion retail across 25 European markets. In 2024, the company served more than 52 million customers and generated €15 billion in gross merchandise volume (GMV) through more than 245 million orders. As a publicly traded company, Zalando commits to achieving specific business objectives, requiring sophisticated pricing strategies to optimize financial performance across its vast product portfolio.
To support these objectives, Zalando Pricing manages price setting for around 600,000 products at any given time across all markets. The primary focus lies in determining optimal discount levels, commonly referred to as red prices in markdown pricing terminology. For instance, when pricing a pair of shoes, the system must decide whether to sell at full price or apply a discount (e.g., at 10\%, 20\%, or higher). Given the enormous scale—hundreds of thousands of products across dozens of markets—the company relies heavily on algorithmic pricing for the vast majority of products.

\paragraph{Algorithmic Pricing: Standard vs.\ Sales Campaign Periods}
During standard business periods (non-sales campaign dates), Zalando's pricing system follows a well-established workflow: products enter the platform at non-discounted prices and undergo periodic re-pricing evaluations. The re-pricing process determines discount levels between 0\% and 70\% for upcoming weeks, primarily driven by risk management considerations. When products do not meet sales expectations, discounting helps mitigate inventory risk and prevents excessive stock accumulation.

This discount optimization employs a sophisticated two-stage algorithmic approach: First, item-level forecasting models predict demand and associated costs for various discount scenarios. These forecasts then feed into an optimization engine that selects the optimal discount level based on predefined business objectives. Both components have been extensively documented in prior work (\cite{li2022large} and~\cite{kunz2023deep}).

\begin{figure}[!h]
  \centering
  \includegraphics[width=.9\textwidth]{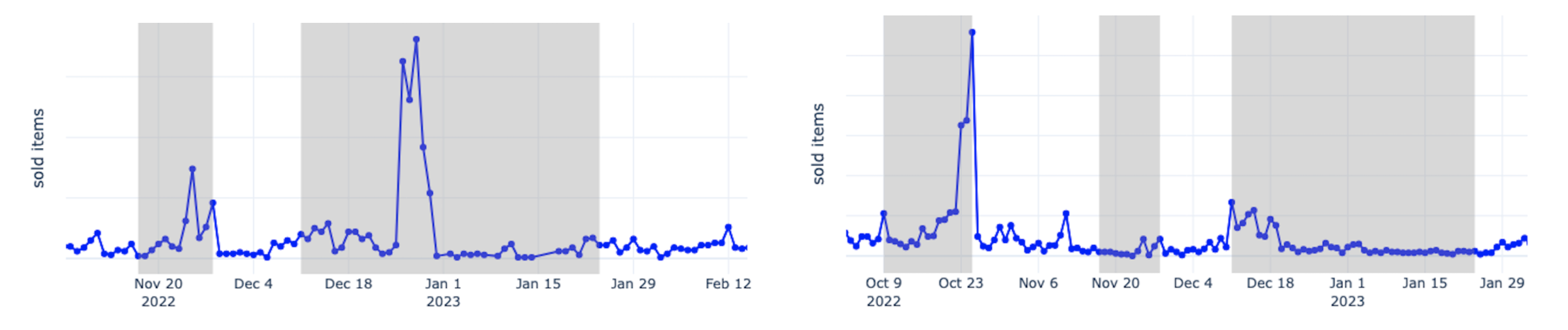}
  \caption{Example articles with daily sales. The gray-shaded areas represent periods with sales events during which we can observe larger fluctuations and particularly large jumps in observed sales compared to the times outside the events.}
  \label{cskus_plot}
\end{figure}

However, sales campaigns, or non-standard business periods, present fundamentally different challenges that have grown increasingly important in recent years. These events, including end-of-season sales, Black Friday, or mid-season promotions, are characterized by high discount levels and more dynamic pricing strategies that require rapid decision cycles with daily or sub-daily price adjustments. The volatile demand patterns exhibit sudden shifts in price elasticity, while irregular event durations and competitive pressure from other retailers necessitate higher intervention frequency from pricing teams. For illustration, Figure~\ref{cskus_plot} shows two example articles and how most of the volatility and larger jumps in demand can be located during sales events.

Around 35\% of each year is taken up by sales events with an average duration of 6.2 days.
Despite their special status, sales events account for a significant amount of Zalando's revenue and profit every year. Discounts are on average 5 percentage points (ppt) higher during sales event periods compared to other days.

\begin{figure}[!h]
  \centering
  \includegraphics[width=.9\textwidth]{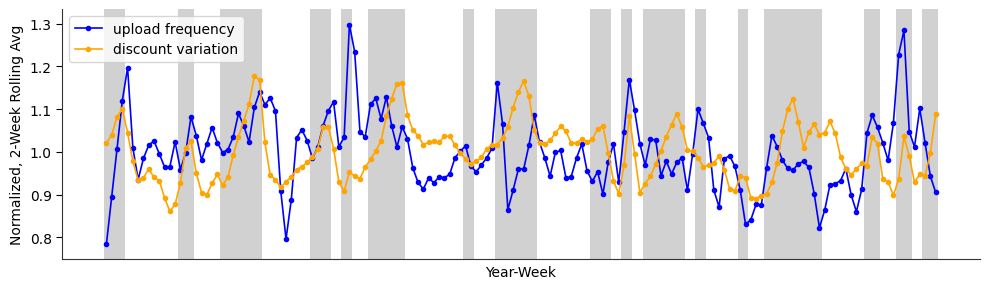}
  \caption{Upload frequency and discount variation (2-week rolling averages, normalized to mean one) since 2023 in Germany. Upload frequency is defined as the ratio of price uploads to the number of products; discount variation is the standard deviation of weekly discounts across the assortment. Periods of sales events are marked in gray. Sales events exhibit both higher upload frequency and greater discount variation, reflecting more frequent price updates and a wider range of offered discounts. Year-week is dropped for anonymity.}
  \label{price_dynamics}
\end{figure}

Figure~\ref{price_dynamics} provides further evidence that sales events require distinct pricing approaches. The figure displays two metrics as 2-week rolling averages, normalized to a mean of one over the observation period: upload frequency (the ratio of price uploads to the number of products) and discount variation (the standard deviation of weekly discounts across the assortment). Periods of sales events are displayed in gray. During sales events, both metrics are consistently elevated—upload frequency increases as more products receive price updates, while discount variation rises as a wider range of discounts is applied across the assortment. These patterns indicate that sales campaigns involve not only more frequent pricing interventions but also greater heterogeneity in discount levels. A weekly-granularity system cannot keep pace with this cadence. The pattern persists across nearly three years of data, underscoring that sales campaigns require a high-frequency pricing solution capable of daily forecast updates and rapid optimization cycles.

These characteristics exposed significant limitations in our existing pricing infrastructure. On the demand forecasting side, our production system relied on a computationally expensive Transformer model operating at weekly granularity with weekly forecast updates. However, sales campaigns require daily resolution to capture rapid demand fluctuations and precise event timing.

The optimization framework was equally problematic. Previous sales campaign pricing relied on a complex combination of time-intensive algorithmic model runs, manual pricing analytics, and heuristic rules incorporating both cost-plus and value-based strategies. Critically, these decisions focused on achieving target average discount levels, with only implicit consideration of key financial performance indicators (KPIs) such as revenue and profit. This approach was clearly suboptimal and motivated a fundamental redesign of the optimization objective to make financial KPIs explicit target variables.

\paragraph{A Tailored Algorithmic Solution for Sales Campaigns}

In this paper, we introduce a specialized demand forecasting and price optimization system that combines time-series forecasting with linear programming optimization in a forecast-then-optimize framework.

\begin{figure}[tbph!]
  \centering
  \includegraphics[width=.9\textwidth]{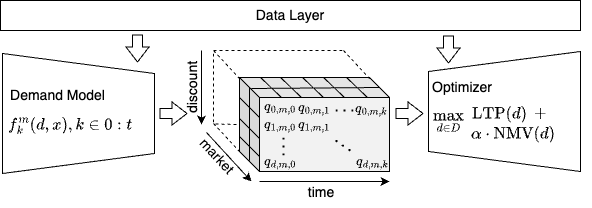}
  \caption{Algorithm Architecture with Optimizer and Forecaster: Demand model $f$ forecasts demand $q$ over a time horizon $0:t$ for market $m$ provided discount $d$ and covariates $x$. The optimizer consumes the persisted data grid in a subsequent stage and selects the optimal discount level $d$ based on the objective function.}
  \label{magpie_architectures}
\end{figure}

Figure~\ref{magpie_architectures} illustrates the proposed algorithm architecture, which consists of three main components connected through a data layer. The demand model generates forecasts for different discount levels and market conditions, producing a discount-time grid that captures sales predictions across various pricing scenarios. This forecasting output feeds into the optimizer, which solves the multi-objective optimization problem by maximizing the weighted combination of long-term profit (LTP) and net merchandise value (NMV) to determine optimal discount levels for each article.

Our solution addresses the unique challenges of sales campaign pricing through several key innovations. First, we implement daily-resolution forecasting using gradient-boosted trees (LightGBM) optimized for sales event dynamics. Second, we develop a multi-objective optimization framework that balances short-term revenue (Net Merchandise Value) with long-term profitability. Third, we achieve computational efficiency that reduces pricing decision time from hours to minutes. Finally, we designed the system as a decision-support tool rather than a fully automated system, maintaining human oversight for financial accountability while providing algorithmic recommendations and trade-off visualizations.

\paragraph{Business Impact}

Through 23 A/B tests across 12 markets during 2023-2024 sales campaigns, we demonstrated that the new pricing system achieves approximately 6\% higher profit while maintaining equivalent performance on sales and revenue metrics compared to the previous manual-algorithmic hybrid approach. Specific markets showed even stronger results, with Germany demonstrating statistically significant improvements of +16.7\% in sales after returns, +17.9\% in net merchandise value, and +16.7\% in profit contribution. Based on these results, the algorithm was successfully deployed to production and now handles the majority of algorithmic pricing decisions for sales campaigns across all markets at Zalando, representing a significant operational improvement in both efficiency and financial performance.

\paragraph{Related Literature}
\label{sec:related_literature}
Pricing is an active, well-studied area with a long tradition in many different disciplines and areas of application (e.g., \citep{You01102003,Kitchener70,Chan2004,pmlr-v202-simchi-levi23a, kumar2022machine, phillips2021pricing}). Here, we focus on the online industry in large-scale scenarios and follow the widely used forecast-then-optimize approach (e.g., \citep{Simchi16,caro12,kedia2020price}).

In particular, we build on previous publications \citep{li2022large, streeck2024tricks}, which described Zalando's algorithmic solution for regular non-sales-event discounting. The existing system employs detailed forecasting models, most notably for demand and return rates, across a 26-week period at weekly granularity spanning 14 countries and covering all 600,000 products. These forecasts, combined with additional initial information such as stock levels, are used to construct a comprehensive model for discount selection. The system considers country-specific sales and stock balancing requirements and can forecast the impact of discount decisions on long-term profitability (LTP). Article-level decisions are linked to market-level objectives through linking constraints, enabling the achievement of specific revenue targets or discount-depth requirements. The existing system applies a Mixed-Integer Programming (MIP) approach with Lagrangian decomposition to handle these constraints, resulting in article-level subproblems that are efficiently solvable as MIPs.

For the optimizer to make effective pricing decisions, it is essential to predict the future impact of these decisions, particularly forecasting future demand under different discount scenarios. Although recent advances in the research community have explored combining prediction and decision-making into a single step \citep{Bertsimas2020, Elmachtoub2022}, a popular approach to markdown pricing publications follow the first-predict-then-optimize framework that we adopt in this work. In this approach, machine learning models leverage historical data to generate inputs for optimization problems before the optimization step. Examples of online markdown pricing following this framework have been outlined in \cite{loh2022promotheus} and \cite{ferreira2016}. A related approach in the airline industry context can be found in \cite{kumar2022machine}, while \cite{phillips2021pricing} provides a general reference on revenue optimization and \cite{reinforcmentlearningforpricing} explore how to utilize reinforcement learning to approach pricing from a game theorist point-of-view.

The most critical component for markdown pricing optimization is the estimation of future demand across different price points. This challenge belongs to the domain of time-series forecasting \citep{petropoulos2022forecasting}. A specialized sub-area focuses on demand forecasting \citep{M5Results, TimDemandForecasting,
Tichy, kunz2023deep}. At Zalando, our production solution for predicting weekly article-level demand during non-sales-event periods is based on a Transformer model outlined in \cite{kunz2023deep}. In contrast, our sales event forecasting approach relies on LightGBM \citep{friedman2001greedy, ke2017lightgbm} as demonstrated in Section \ref{subsection:sales_forecast}. The application of LightGBM has proven successful in the M5 competition for forecasting Walmart sales, where it outperformed all other methods, including deep-neural-network-based architectures \citep{M5Results, M5UncertaintyResults}.

Since we need to estimate price-dependent demand, our problem extends beyond classical prediction and enters the domain of causal inference. The goal becomes estimating the causal effect of a treatment (price change) on an outcome (demand). \cite{schultz2024causal} investigate a Double Machine Learning approach applied to both simulated and real Zalando sales data for this purpose.

As we provide forecasts at both weekly and daily levels, the question arises whether hierarchical forecasting along the temporal dimension could improve predictions. Preliminary investigations \cite{prato_talk2023} indicate that reconciling daily with weekly predictions presents non-trivial challenges and remains subject to future work. For another successful application of hierarchical forecasting to demand forecasting, see \cite{Sprangers2024} or \cite{Petropoulos12122025}.

Finally, \cite{Huelden2024} provides comprehensive details on pricing experimentation processes and methodologies at Zalando. These frameworks are used to quantify the causal effects of our algorithmic pricing improvements.

The paper proceeds as follows: Section \ref{section:pricing_algorithm} details the pricing algorithm architecture and multi-objective optimization framework, Section \ref{subsection:sales_forecast} presents the demand forecasting methodology, Section \ref{app:simulated_data} analyzes training loss functions through simulation, and Section \ref{subsec:experiment} reports experimental validation and business results.

\section{Sales Events Pricing Algorithm}
\label{section:pricing_algorithm}
This section presents the theoretical foundation and technical framework of our sales events pricing algorithm. We begin in Section \ref{subsection:objective} by modeling the business objective that drives our optimization approach. Specifically, we address Zalando's pricing requirements that necessitate balancing short-term business goals, such as revenue optimization during sales events, with long-term objectives, particularly profit maximization. Section \ref{subsection:magpie_model} then provides a comprehensive model description that connects the decision variables to our optimization objective, including the derivation of sales dynamics during sales events and the optimization algorithm for determining optimal prices given forecasting inputs.
Since the optimizer requires estimates for future revenue and long-term profit, it depends on several forecasting components that collectively enable the calculation of these business objectives. Most critically, we need accurate sales forecasts across different discount levels, which is addressed in detail in Section \ref{subsection:sales_forecast}.

\subsection{Pricing Objective}
\label{subsection:objective}
Our pricing solution aims to provide Zalando stakeholders with a decision-support tool that recommends discounts aligned with their business strategy while showcasing alternative options with potentially superior characteristics. More specifically, given an assortment of articles, the solution recommends an optimal discount for each article over a future time frame (such as a sales event) that maximizes a carefully designed objective function. We refer to such assortment-level discount recommendations as \emph{pricing offers}.
To enable meaningful comparison between different offers, we identified three major drivers of sales events through stakeholder consultation:

\begin{enumerate}
  \item\label{Driver1} Revenue optimization: Maximize top-line performance obtained from sales during the sales event, measured as Net Merchandise Value (NMV),
  \item\label{Driver2} Customer engagement: Drive customer acquisition and retention
  \item\label{Driver3} Inventory management: Clear overstocked items in a profitable manner.
\end{enumerate}

Since our approach focuses on article-centric rather than customer-centric decisions, we address customer acquisition and retention indirectly through NMV optimization, effectively combining objectives (\ref{Driver1}) and (\ref{Driver2}).

However, revenue optimization can conflict with (\ref{Driver3}). Optimizing for short-term revenue typically leads to higher discounts and faster stock clearance, while optimizing for profit may indicate that selling an article at a later time with a more profitable price is preferable. This tension necessitated providing stakeholders with the flexibility to balance short-term and long-term interests, including the ability to make manual interventions during unforeseen circumstances such as the COVID-19 pandemic in 2020.

While NMV is straightforward to model, we use long-term profitability (LTP) as a proxy for effective stock management. At a high level, LTP incorporates profit generated over the next season together with potential revenue from post-season sales, as detailed in Section \ref{subsection:magpie_model}.

This scenario represents a classic multi-objective optimization problem. We employ the standard approach of identifying Pareto-efficient offers \citep{Ehrgott2005}—offers that cannot be dominated by competing alternatives in both NMV and LTP dimensions. However, since many Pareto-efficient offers typically exist, expert stakeholder knowledge is essential for final selection.

To enable this, we use the \emph{Weighted Sum Method} \citep{Ehrgott2005}.
For our concrete setting, it is formulated as follows:
For some $\alpha \in [0, \infty)$, for articles $i \in N$, and discounts $d_i \in D$, our optimization objective is
\begin{equation}\label{eq:optProblem}
  \max_{d_{i}\in D} \ \text{LTP}_i(d_i) + \alpha \cdot \text{NMV}_i(d_i)  \ \forall
  i
  \in N.
\end{equation}
As established in optimization theory, any solution maximizing this objective will be Pareto-efficient \citep{marler2010weighted}.

\begin{figure}[tbph!]
  \centering
  \includegraphics[width=.75\textwidth]{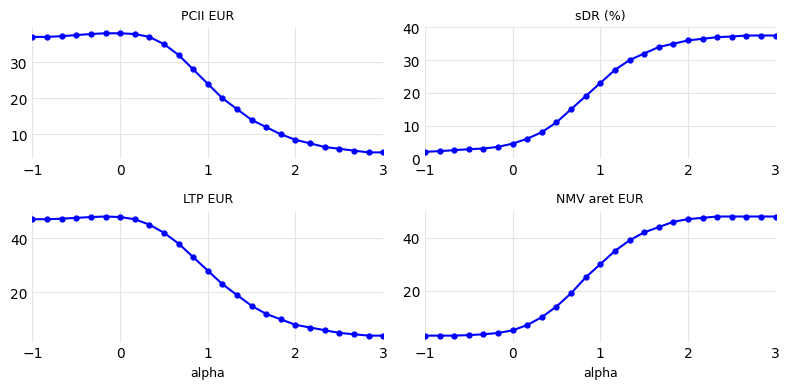}
  \caption{Sampling allows us to visualize trade-offs between short-term and long-term KPIs as well as the overall recommended discount level. The sales discount rate (sDR) is given by 1 - (GMV after discount/GMV before discount).}
  \label{fig:magpie_properties}
\end{figure}

Stakeholders can influence the balance between short-term and long-term metrics by selecting appropriate values of $\alpha$. In practice, we compute offers for multiple $\alpha$ values, enabling stakeholders to evaluate options using not only predicted NMV and LTP but also various other offer characteristics, as illustrated in Figure~\ref{fig:magpie_properties}. This figure demonstrates how different $\alpha$ values affect key performance indicators across an example assortment. The four panels show the trade-offs between profit contribution (PCII), sales discount rate (sDR, i.e., sales-weighted average discounts), long-term profit (LTP), and net merchandise value (NMV) as $\alpha$ varies. As $\alpha$ increases (prioritizing short-term revenue), we observe higher discount rates and NMV but diminishing long-term profitability, clearly illustrating the fundamental tension between short-term and long-term objectives that our multi-objective framework addresses.

The parameter $\alpha$ has an intuitive interpretation as the trade-off rate between LTP and NMV. To understand this, consider a scenario where discounts are chosen from a continuous interval for a single article, with both LTP and NMV being everywhere differentiable functions of discount.
Then
\begin{equation}\label{eq:alpha}
  -\frac{\text{NMV}'(d)}{\text{LTP}'(d)} = \alpha^{-1}.
 \end{equation}
This relationship follows from first-order optimality conditions. At the margin, our objective requires choosing discounts such that the return from increasing discount $d$ weighted by $\alpha \cdot \text{NMV}'(d)$ equals the loss in long-term profit $-\text{LTP}'(d)$. Therefore, $\alpha$ quantifies how much LTP will be sacrificed to increase NMV.

\subsection{Model Description}
\label{subsection:magpie_model}

After establishing our optimization objective, we proceed to derive the components needed to compute equation \eqref{eq:optProblem}, enabling us to describe the complete optimization procedure.

The foundation for computing \eqref{eq:optProblem} is predicting article sales within the sales event across different price points, denoted as $S_{i, d}^{t}$. This requires a sophisticated forecasting approach detailed in Section \ref{subsection:sales_forecast}. Therefore, we assume in the following
that for a given article $i$, discount $d$ and for each individual day $t \in
T$ of the sales event this quantity is $S_{i, d}^{t}$. In the
following we will assume one fixed article, such that we drop the index~$i$ for ease of notation.

Since our forecaster operates without knowledge of future stock situations, we first aggregate sales forecasts $S^t_d$ over the planned sales event duration and constrain them by current stock levels $M$ to calculate final sales projections $s_d$:
\begin{equation}
  s_d = \min(M, \sum_{t \in T} S_{d}^{t}).
\end{equation}
Note that sales predictions are discount-dependent. The reduced price under discount~$d$ is given by $(1-d) \cdot
P$, where $P$ represents the article's undiscounted price.

The primary cost consideration for online fashion articles is customer returns. Therefore, Zalando's pricing system incorporates proven methodology for predicting article-dependent return rates $R$, representing the fraction of sold articles that customers return. Combined with country-dependent value-added tax (VAT), we calculate NMV as:
\begin{equation}
  \text{NMV}(d) = (1-R) \cdot \frac{(1-d) \cdot s_d \cdot P}{(1 + \text{VAT})}.
\end{equation}

To derive $\text{LTP}(d)$, we compute both in-event short-term profit and future profit potential. For short-term profit calculation, we utilize cost forecasts $C$ employed in other Zalando pricing applications, which determine return costs and fulfillment costs through conditional averages over historical data:
\begin{equation}\label{eq:p}
  p(d) = \text{NMV}(d) - C \cdot s_d.
\end{equation}
For future profit opportunities, we add opportunity cost $\phi(s_d)$. This future profit potential is inherently complex, depending on factors including stock level $M$, sales $s_d$, future discounting policies, and potential responses to sales event outcomes.

In our implementation, we leverage results from our weekly non-sales-event pricing system (see Section~\ref{sec:related_literature}) to extract forecasts for $\phi(s_d)$. For simplicity, consider that the existing system estimates $p(w, \pi)$—the profit an article will generate over the next $w
\in W = \{0, \ldots, 25\}$ weeks under a long-term profit optimal policy. We then calculate the per-unit profit average $\gamma$ using the total future inventory $m$:
\begin{equation}
  \gamma \cdot m = \sum_{w > 0} p(w).
\end{equation}

This enables us to calculate $\text{LTP}(d)$ for sales events as:
\begin{equation}
  \text{LTP}(d) = p(d) + (M - s_d (1-R)) \cdot \gamma.
\end{equation}

While this approach of using a single constant $\gamma$ cannot adapt future discounting policies in response to sales outcomes, it possesses valuable qualities for our setup: it computes quickly from existing data and aligns well with the weekly results from the existing system in terms of both absolute $\gamma$ values and relative costs of discount adjustments near system-selected levels.

Our modeling of LTP and NMV assumes no interdependencies between articles—discounting one article does not influence sales projections of others. Similarly, objective \eqref{eq:optProblem} does not link different articles. Therefore, we can compute NMV and LTP individually for each article and discount combination to obtain objective values, then determine optimal discounts by sorting these values per article.
This approach enables massively parallel computation on large clusters, solving the assortment discounting problem in minutes rather than the hours required by the existing system, and at a fraction of the computational cost.

\section{Forecasting demand for sales events}
\label{subsection:sales_forecast}

As we have outlined before, the optimization problem Eq.~\eqref{eq:optProblem}
becomes deterministic once we have fixed its inputs. Towards that end, we need
to provide forecasts $f^{m}(d, x)$ that predict the amount of sales for a
single article in a given market $m$ at a given discount $d$ and a set of
relevant features $x$. Since Eq.~\eqref{eq:optProblem} is agnostic to the
specific forecasting methods used and only cares about the final predictions,
we can easily plug-and-play with different types of forecasts. The only
requirement these should fulfill is to be monotonically increasing in the discount level~$d$.

\subsection{Solution and Methods}\label{subsec:ForecastingMethod}
For each market $m$ and each day $k$ within our forecast horizon, we train a
model $f_{k}^{m}(d, x)$, to predict the amount of sales for a single
stock keeping unit (SKU) at a given price level and a set of relevant features
$\mathcal{X}$.\footnote{In the following discussion, we use "price level"
interchangeably with the term "discount," i.e., the relative price level w.r.t.
some initial price}. Please also see Table~\ref{tab:example-table} for further
details on our set of features.
\begin{table}[htbp]
\centering
\begin{tabular}{llc}
\toprule
Feature & Description & Dimension (Type) \\
\midrule
Past observed sales & Past sales over last \pasthorizon days & \pasthorizon D (num) \\
Past observed stock & Stock level over last \pasthorizon days & \pasthorizon D (num) \\
Past views & Past webpage views of the article & \pasthorizon D (num) \\
Past discount & Discount of last \pasthorizon days & 1D (num) \\
Commodity group 1--5 & Category/type of article & 5 $\times$ 1D (cat) \\
Article brand & Brand of each article & 1D (cat) \\
Black price & Non-discounted price & 1D (num) \\
Future discount & Observed/set discount & 1D (num) \\
Future voucher event & Voucher event indicator & 12D (binary) \\
Future sales event & Sales event indicator & 12D (binary) \\
Holiday & Holiday flag & 1D (binary) \\
Day of month & The day number of the month & 1D (num) \\
Day of week & Weekday & 1D (num) \\
\bottomrule
\end{tabular}
\caption{Features used with the demand forecast.}
\label{tab:example-table}
\end{table}

When training a model, be it for the purpose of experimentation or for
production, we use data only from relevant sales events in the past and the test
sets typically cover a single sales event. We discard 90\% of articles at random that did not have a sale in the past 10 days to better balance our training data.
We also enforce a monotonicity constraint on the future discount feature, such that a
higher discount cannot lead to lower demand predictions. By incorporating such
business logic we also drastically reduce the amount of possible solutions
for the downstream optimizer.

\subsection{Choosing the best forecasting model}
We compared five different forecasting models:
\begin{description}
    \item[Naive Model:] A model using exponential smoothing and an economic model for elasticity,
    \begin{equation}
        \log(f_{k}^{m}(d^*, x)) = \log(\lambda\sum_{i=1}^{10} (1-\lambda)^{i-1} \text{sales}_{t-i}) + \alpha(\log(1-d^*) - \log(1-d_{t-1}))
    \end{equation}
    where $\lambda \in (0, 1]$ is a smoothing parameter and $\alpha$ denotes the elasticity.
    \item[Weekly Model:] Disaggregating our Transformer model \citep{kunz2023deep} to daily resolution by using a learned distribution over the weekdays.
    \item[Gradient Boosted Trees:] Gradient-boosted regression tree, where we use the implementation in LightGBM \citep{friedman2001greedy, ke2017lightgbm}.
    \item[TSMixer:] Implementation of the TSMixer model \citep{chen2023tsmixerallmlparchitecturetime}.
    \item[MLP:] A simple feed-forward neural network.
\end{description}

For evaluating our model, we predict the amount of sales for the materialized
discount $d^*$ and compare it to the observed number of sales. As an accuracy
metric, we calculate the so-called sales error for each day and each market:
\begin{equation}
  \text{sales error}_{k}^{m} = \sqrt{\frac{\sum \text{black price}(\cdot
    f_{k}^{m}(d^*, x) - \text{sales})^2}{\sum \text{black price} \cdot
      \text{sales}^2}} .
\end{equation}

In addition, we compare the predicted with the realized GMV to measure our
model’s bias:
\begin{equation}
  \hat{\text{gmv}}_{k}^{m} = \sum_{csku} \text{black price} \cdot (1 - d^*) \cdot
  f_{k}^{m}(d^*, x),
\end{equation}
\begin{equation}
  \text{gmv error}_{k}^{m} = \left|\frac{\hat{\text{gmv}}_{k}^{m} -
    \text{gmv}_{k}^{m}}{\text{gmv}_{k}^{m}}\right|.
\end{equation}

Apart from the \textit{Weekly Model}, we hyperparameter tuned all other models on a validation set to minimize
the sales error. We were able to tune the \textit{Naive Model} and
\textit{Gradient Boosted Trees} more extensively than the neural networks given that \textit{Naive Model} is simple and that \textit{Gradient Boosted Trees} has a very effective and mature implementation in LightGBM which allowed us to train them very fast. This was identified in \cite{Januschowski2022} for the effectiveness of tree-based forecasting algorithms.

The biggest impact we saw came from using appropriate loss functions, which in case of \textit{Gradient Boosted Trees} means using Tweedie loss and in case of the two neural networks, the negative binomial loss function.

Comparing the results (see Table~\ref{table:emp-results}) we see a relatively close performance between the three ML models (\textit{Gradient Boosted Trees}, \textit{TSMixer}, \textit{MLP}) and a significant gap to the \textit{Naive Model} and our approach to reusing our \textit{Weekly model} to predict daily sales by disaggregating. On Demand Error and GMV Error, our main internal KPIs for backtesting, the Gradient Boosted Trees perform the best. On the RMSE and MAPE we see a very close performance between the two neural networks. Especially the relatively simple \textit{MLP} delivers the best performance on RMSE and MAPE which, in contrast to the Demand Error, is uniformly weighted over our assortment.

Due to these similar results, all three ML models are suitable candidates. We chose to use \textit{Gradient Boosted Trees} for the final solution. On top of stellar performance, they provide the fastest and most stable implementation in LightGBM, providing a training time of 7h compared to 24h for MLP and 18h for the TSMixer model.

\begin{table}[htbp]
\centering
\begin{tabular}{lcccccccc}
\toprule
 & \multicolumn{2}{c}{Demand Error} & \multicolumn{2}{c}{GMV Error} & \multicolumn{2}{c}{RMSE} & \multicolumn{2}{c}{MAPE} \\
\cmidrule(lr){2-3} \cmidrule(lr){4-5} \cmidrule(lr){6-7} \cmidrule(lr){8-9}
Forecaster & Mean & Std & Mean & Std & Mean & Std & Mean & Std \\
\midrule
Naive & 0.743 & 0.288 & 0.256 & 0.243 & 4.11 & 3.55 & 0.059 & 0.015 \\
Weekly & 0.735 & 0.197 & 0.300 & 0.289 & 4.17 & 3.72 & 0.061 & 0.017 \\
GBT & \textbf{0.574} & 0.105 & \textbf{0.142} & 0.129 & 3.24 & 2.89 & 0.052 & 0.014 \\
TSMixer & 0.614 & 0.105 & 0.187 & 0.138 & 3.12 & 2.95 & 0.042 & 0.009 \\
MLP & 0.612 & 0.111 & 0.154 & 0.116 & \textbf{3.02} & 2.60 & \textbf{0.040} & 0.010 \\
\bottomrule
\end{tabular}
\caption{Comparison of the candidate models. The metrics are calculated over 212 different days between 2024-05-01 and 2024-12-01 spanning half a year and capturing at least one of each major sales event. We report the mean and standard deviation over the 212 days.}
\label{table:emp-results}
\end{table}

\section{Training Loss Comparison using Simulated Data}
\label{app:simulated_data}
Observing a significant impact of our training loss choice in backtesting, we were
also interested in exploring its impact on our pricing algorithm as a whole. We therefore
conducted a simulation study comparing the effect of using different loss functions in
combination with the most promising model candidate
\textit{Gradient Boosted Trees}.

\subsection{Simulation setup}

For the simulation study, we use a modified setup of
\cite{schultz2024causal}.
The main modification is that, rather than using a linear price response, we employ a
constant-elasticity price response:
\begin{equation}
  q_{it}(d) = q_{it}(0) (1 - d)^{-\epsilon_i},
\end{equation}
where $q_{it}(d)$ is the demand for a given article $i$ at time $t$ and
discount $d$ and $\epsilon_i$ is the
elasticity of demand for article $i$ (by convention, our elasticities
are positive).

As described in \cite{schultz2024causal}, the black-price demand
$q_{it}(0)$ contains a trend, seasonality and
noise components.
In contrast to \cite{schultz2024causal}, we generate the noise term in $q_{it}$ from
a log-normal
distribution (as opposed to the normal distribution).
This allows for a more realistic demand behavior, using the constant-elasticity price response
(e.g.\ we practically eliminated the problem of negative demand appearing,
which required dropping some articles
before).
Moreover, the elasticities $\epsilon_i$ are generated from a log-normal
distribution.

As described in \cite{schultz2024causal}, the simulation starts with a
given level of stock for each article and
uses a simple discounting heuristic attempting to clear the stock before the
end of the simulation,
i.e., after 100 time periods ("weeks").
The possible discounts are between 0 and 50\% in 10\% increments.

For each week, the simulation outputs the materialized sales and discounts, e.g.
for any given week, it outputs the full grid of sales for all possible
discounts
(assuming that the previous weeks had the materialized discounts and sales).

\subsection{Comparison of MSE and Tweedie training losses}

To test our pricing algorithm described in Section~\ref{section:pricing_algorithm} on the simulated data, we use the first $X$ weeks
for training
(we choose $X=50$ for the experiments presented in this section), the week
$X+1$ for prediction by a forecast model,
and the remaining weeks to estimate the per-unit profit average $\gamma$ for
running our optimization on the simulated dataset.
Computing profit in our optimizer requires setting several additional parameters
(such as fulfillment costs, after-season
residual value etc.).
We calibrate these parameters to our Zalando data.

Now, we are equipped to train the models.
For this experiment, we train two tree forecasting ("S-Learner") models,
sharing common features.
The only difference between the models is the loss function used for training.
The first model uses the mean squared error (MSE) loss, while the other uses
the Tweedie loss, just like the forecast
model for the real pricing algorithm.

We compute the forecast accuracy metrics for week $X+1$ and obtain that the two
models have very similar performance
(Tweedie model has a sales error lower by $1.8\%$ than the MSE model, and a
slightly higher GMV error).

Analyzing the local (arc) elasticities for each article does not
immediately reveal which model performs better.
For computing the arc elasticity, we select a point around the midpoint of the
grid, $d=30\%$.
To avoid the impact of outliers, we look at the median elasticity, and find
that the MSE model's median is 9\% lower
than the ground truth median, while the Tweedie model overestimates the median
by 8\%.

Next, we run the two models through the optimizer.
As a benchmark, we also run optimization on an "oracle" forecast model, that
uses the oracle sales grid in week
$X+1$ (but uses the same $\gamma$ coefficient as the other forecast models).
The results for the profit-optimal ($\alpha=0$) setting are presented in
Table~\ref{tab:sim_magpie_kpis}.
We see that the MSE model tends to "over-forecast" and "under-deliver" (it
promises even better results than the oracle
model, but delivers well below it).
On the other hand, the Tweedie model is more conservative in its forecasts, but
delivers on them much better.
Therefore, we conclude from the simulation that the forecast model trained with
Tweedie loss is better suited for the
pricing optimization use case.

\begin{table}[tbp!]
\centering
    \input{sim_magpie_kpis}

  \caption{Relative difference of forecasted and materialized KPIs with respect to the oracle model in the simulation study.}
\label{tab:sim_magpie_kpis}

\end{table}

\section{Experiments and Results}%
\label{subsec:experiment}
This section presents our experimental validation of the pricing algorithm in real-world conditions. We first describe our A/B testing methodology, including the challenges of implementing controlled experiments across multiple markets during high-stakes sales campaigns. We then analyze performance outcomes, comparing our algorithmic approach against the existing manual pricing process. Beyond demonstrating algorithmic performance, this validation was crucial for building stakeholder confidence in deploying machine learning for critical pricing decisions.
\subsection{Experimentation planning \& setup}

Testing presented several significant challenges that required careful experimental design. First, we had to navigate the complex human-versus-machine comparison, contrasting our new algorithmic discounts with the status quo pricing process that relied on partly manual discount derivations. This comparison required close alignment with pricing managers who had developed expertise and intuition over years of manual pricing decisions. Establishing trust and buy-in from these domain experts was essential for successful experimentation.

Second, testing opportunities were inherently limited, externally determined, and time-constrained since our algorithm is specifically designed for sales events. Unlike continuous optimization problems, we could only validate performance during discrete, strategically important sales periods when pricing errors could have substantial financial consequences. This constraint required us to maximize the value of each testing opportunity while ensuring robust statistical inference.

Third, the high-stakes nature of sales events—which can represent significant portions of quarterly revenue—necessitated extremely careful risk management. Any algorithmic failure during major sales periods could result in substantial financial losses and damage to customer relationships.

The primary success criterion for our testing efforts was demonstrating that the new pricing system performs at least on par with our current pricing solution (manual modifications applied to discounts from the existing system) across key commercial outcomes such as NMV and profitability. This "do no harm" threshold represented a sufficient justification for adoption of the new pricing algorithm, given that it dramatically reduces operational complexity in our pricing processes.

For experimental design, articles participating in A/B tests were divided into two equally-sized groups: those continuing with the previous method of manually derived discounts (control group) and those receiving algorithmically-generated discounts (treatment group). Importantly, we implemented article-level rather than customer-level pricing experiments to avoid price discrimination and maintain ethical pricing practices.

As we employed article-level randomization, we had to ensure that treatment and control groups represented similar articles in terms of product characteristics as well as historical profit and NMV contribution under varying price levels. To achieve this balance, we employed a clustered randomization scheme designed to minimize potential substitution effects between similar products while balancing selected pre-intervention variables during the randomization step. Our standard experimentation methodology is described in greater detail in \cite{Huelden2024}.

\subsection{A/B test results \& decision making}
To validate our proposed pricing algorithm and quantify its commercial impact, we conducted a total of 23 A/B tests across twelve European markets: Belgium, Denmark, Finland, France, Germany, Italy, the Netherlands, Norway, Poland, Spain, Sweden, and Switzerland. The tests were run in multiple waves throughout 2023 and 2024 during major sales campaigns, including end-of-season sales (EOSS) and mid-season sales (MSS). Table~\ref{tab:experiment_summary} provides an overview of the experimental setup. Experiment duration ranged from 4 to 10 days, with an average of 7 days. Sample sizes varied from approximately 200,000 to over 1.8 million articles per experiment, covering 6.2 million articles in total.

\begin{table}[htbp]
\centering
\begin{tabular}{lllrr}
\toprule
Market & Start Date & End Date & Days & Articles \\
\midrule
IT, NL & 2023-10-09 & 2023-10-15 & 6 & 568,458 \\
NL & 2024-01-24 & 2024-01-28 & 4 & 202,326 \\
NL & 2024-01-24 & 2024-01-28 & 4 & 201,768 \\
IT & 2024-02-01 & 2024-02-11 & 10 & 201,109 \\
IT & 2024-02-01 & 2024-02-11 & 10 & 201,250 \\
ES, IT, NL & 2024-04-08 & 2024-04-14 & 6 & 1,083,970 \\
BE & 2024-04-11 & 2024-04-21 & 10 & 361,633 \\
CH & 2024-04-14 & 2024-04-21 & 7 & 368,420 \\
DE & 2024-04-28 & 2024-05-05 & 7 & 363,146 \\
DK, FI, NO, PL, SE & 2024-10-05 & 2024-10-16 & 9 & 820,960 \\
BE, DE, ES, FR, IT, NL & 2024-10-13 & 2024-10-23 & 9 & 1,831,587 \\
\midrule
Total/Avg. & & & 7 & 6,204,627 \\
\bottomrule
\end{tabular}
\caption{Summary statistics by experiment wave. Each row represents a distinct experiment wave; rows with multiple markets indicate concurrent A/B tests run in parallel, contributing to the total of 23 A/B tests.}
\label{tab:experiment_summary}
\end{table}

We analyzed our experiments using the difference-in-differences methodology \citep{abadie2005} to compute uplift in profit contribution (PCII), net merchandise value (NMV), and sales after returns (SIAR). This approach accounts for temporal trends and market-level factors that could confound the comparison between algorithmic and manual pricing approaches.

Table~\ref{tab:ab_test_results} summarizes the aggregated outcomes across all experiments, showing the overall effect sizes and confidence intervals.
Combining results across all test waves through meta-analysis using a Bayesian hierarchical model, we observed a statistically significant overall increase of 6\% in PCII and an insignificant increase in NMV. SIAR were not statistically meaningfully affected. The Bayesian approach allowed us to properly account for heterogeneity across different markets and time periods while providing robust uncertainty quantification around our meta study effect estimates.

\begin{table}[htb]
\centering
\input{ab_test_results}
\caption{Meta-analysis results of algorithmic pricing performance across A/B tests.}
\label{tab:ab_test_results}
\end{table}

Further experiments in larger markets such as Switzerland (CH) and Germany (DE) provided additional validation and highlighted the potential for even stronger commercial uplift in high-volume markets. Germany, representing one of our most significant markets, demonstrated statistically significant improvements of +16.7\% for SIAR, +17.9\% for NMV, and +16.7\% for PCII. These results suggest that algorithmic pricing may be particularly effective in markets with sufficient scale and data richness to enable more sophisticated optimization.

The 6\% profit improvement represents substantial value creation when scaled across Zalando's sales event portfolio. This improvement stems from the algorithm's ability to better balance short-term revenue objectives with long-term profitability considerations, as well as its capacity to process vast amounts of historical data to identify optimal pricing patterns that may not be apparent to human decision-makers. The strong performance in major markets is especially encouraging because these represent the highest-stakes environments where pricing decisions have the greatest absolute impact on business performance. The consistency of positive results across multiple metrics (sales, revenue, and profit) indicates that the algorithm successfully navigates the multi-objective optimization challenge without simply optimizing one metric at the expense of others.

Based on these results, the algorithm was deployed to production and now handles the majority of algorithmic pricing decisions for sales campaigns across the company. This deployment represents a significant operational transformation, enabling pricing teams to focus on strategic oversight rather than routine discount calculations. The scalability and efficiency gains have been particularly valuable during peak sales periods when rapid pricing decisions across hundreds of thousands of products are critical for business success.

\section{Discussion and Conclusion}

This paper describes the successful development, validation, and deployment of a specialized algorithmic pricing system designed for sales campaigns in fashion e-commerce. Our work addresses the unique challenges of sales event pricing, which differ fundamentally from standard business periods due to their dynamic nature, compressed timelines, and conflicting business objectives.

Our approach combines daily-resolution forecasting using gradient-boosted trees with a multi-objective optimization framework that balances short-term revenue objectives with long-term profitability. The forecast-then-optimize architecture enables scalable parallel computation while maintaining flexibility for stakeholder input through Pareto-efficient solution selection.
Beyond technical contributions, this work illustrates the complete journey from research prototype to production deployment. Our experience emphasizes critical factors for algorithmic adoption: clearly defined performance guardrails, gradual testing with "do no harm" objectives, extensive stakeholder communication, and building trust through transparent experimentation.

Several opportunities for improvement remain. The current long-term profitability calculation could be enhanced through more sophisticated modeling approaches. Articles with minimal historical sales present forecasting challenges that warrant specialized treatment. Important business constraints such as voucher integration require further development. Additionally, factors like demand uncertainty and cross-price elasticity are not explicitly modeled, representing important research directions.

The 6\% profit improvement and successful production deployment provide compelling evidence for the value of algorithmic approaches in dynamic pricing environments. Our experience offers valuable insights for bridging the gap between academic research and commercial applications, emphasizing that technical excellence must be coupled with careful experimental validation and stakeholder engagement to achieve real-world impact.

\section{Disclosure Statement}
All authors but Tim Januschowski were working at Zalando SE while the solution was developed and during the creation of this paper. Tim Januschowski has also been working at Zalando SE while the solution was developed and was working at Databricks during the creation of this paper.

\clearpage

\bibliographystyle{unsrtnat}
\bibliography{references}

\end{document}

%% file: sim_magpie_kpis.tex
\begin{tabular}{lccc}
\toprule
Model & LTP & NMV & PCII \\
\midrule
MSE forecasted & 100.05\% & 107.31\% & 115.02\% \\
MSE materialized & 99.96\% & 88.54\% & 93.07\% \\
Tweedie forecasted & 99.97\% & 91.60\% & 94.21\% \\
Tweedie materialized & 99.97\% & 91.02\% & 91.57\% \\
\bottomrule
\end{tabular}

%% file: ab_test_results.tex
\begin{tabular}{lcc}
\toprule
KPI & Effect & 95\% CI \\
\midrule
PCII & 6.00\% & [0.79\%, 11.03\%] \\
NMV & 2.23\% & [$-$0.83\%, 5.33\%] \\
SIAR & 0.56\% & [$-$3.10\%, 4.26\%] \\
\bottomrule
\end{tabular}

%% file: main.bbl
\begin{thebibliography}{34}
\providecommand{\natexlab}[1]{#1}
\providecommand{\url}[1]{\texttt{#1}}
\expandafter\ifx\csname urlstyle\endcsname\relax
  \providecommand{\doi}[1]{doi: #1}\else
  \providecommand{\doi}{doi: \begingroup \urlstyle{rm}\Url}\fi

\bibitem[Li et~al.(2022)Li, Simchi-Levi, Sun, Wu, Fux, Gellert, Greiner, and
  Taverna]{li2022large}
Hanwei Li, David Simchi-Levi, Rui Sun, Michelle~Xiao Wu, Vladimir Fux, Torsten
  Gellert, Thorsten Greiner, and Andrea Taverna.
\newblock Large-scale price optimization for an online fashion retailer.
\newblock In Volodymyr Babich et~al., editors, \emph{Innovative Technology at
  the Interface of Finance and Operations: Volume II}, pages 191--224. Springer
  International Publishing, 2022.
\newblock \doi{10.1007/978-3-030-81945-3_8}.

\bibitem[Kunz et~al.(2023)Kunz, Birr, Raslan, Ma, and
  Januschowski]{kunz2023deep}
Manuel Kunz, Stefan Birr, Mones Raslan, Lei Ma, and Tim Januschowski.
\newblock Deep learning based forecasting: a case study from the online fashion
  industry.
\newblock In \emph{Forecasting with artificial intelligence: theory and
  applications}, pages 279--311. Springer, 2023.

\bibitem[You(2003)]{You01102003}
P-S You.
\newblock Dynamic pricing of inventory with cancellation demand.
\newblock \emph{Journal of the Operational Research Society}, 54\penalty0
  (10):\penalty0 1093--1101, 2003.
\newblock \doi{10.1057/palgrave.jors.2601619}.

\bibitem[Kitchener(1970)]{Kitchener70}
A.~Kitchener.
\newblock Pricing strategy.
\newblock \emph{Journal of the Operational Research Society}, page 487, 1970.

\bibitem[Chan et~al.(2004)Chan, Shen, Simchi-Levi, and Swann]{Chan2004}
L.~M.~A. Chan, Z.~J.~Max Shen, David Simchi-Levi, and Julie~L. Swann.
\newblock \emph{Coordination of Pricing and Inventory Decisions: A Survey and
  Classification}, pages 335--392.
\newblock Springer US, Boston, MA, 2004.
\newblock ISBN 978-1-4020-7953-5.
\newblock \doi{10.1007/978-1-4020-7953-5_9}.
\newblock URL \url{https://doi.org/10.1007/978-1-4020-7953-5_9}.

\bibitem[Simchi-Levi and Wang(2023)]{pmlr-v202-simchi-levi23a}
David Simchi-Levi and Chonghuan Wang.
\newblock Pricing experimental design: Causal effect, expected revenue and tail
  risk.
\newblock In Andreas Krause, Emma Brunskill, Kyunghyun Cho, Barbara Engelhardt,
  Sivan Sabato, and Jonathan Scarlett, editors, \emph{Proceedings of the 40th
  International Conference on Machine Learning}, volume 202 of
  \emph{Proceedings of Machine Learning Research}, pages 31788--31799. PMLR,
  23--29 Jul 2023.
\newblock URL \url{https://proceedings.mlr.press/v202/simchi-levi23a.html}.

\bibitem[Kumar et~al.(2022)Kumar, Boluki, Isler, Rauch, and
  Walczak]{kumar2022machine}
Ravi Kumar, Shahin Boluki, Karl Isler, Jonas Rauch, and Darius Walczak.
\newblock Machine learning based framework for robust price-sensitivity
  estimation with application to airline pricing, 2022.

\bibitem[Phillips(2021)]{phillips2021pricing}
Robert~L Phillips.
\newblock \emph{Pricing and revenue optimization}.
\newblock Stanford university press, 2021.

\bibitem[Ferreira et~al.(2016{\natexlab{a}})Ferreira, Lee, and
  Simchi-Levi]{Simchi16}
Kris~Johnson Ferreira, Bin Hong~Alex Lee, and David Simchi-Levi.
\newblock Analytics for an online retailer: Demand forecasting and price
  optimization.
\newblock \emph{Manufacturing \& Service Operations Management}, 18\penalty0
  (1):\penalty0 69–88, February 2016{\natexlab{a}}.
\newblock ISSN 1526-5498.
\newblock \doi{10.1287/msom.2015.0561}.
\newblock URL \url{https://doi.org/10.1287/msom.2015.0561}.

\bibitem[Caro and Gallien(2012)]{caro12}
Felipe Caro and J\'{e}r\'{e}mie Gallien.
\newblock Clearance pricing optimization for a fast-fashion retailer.
\newblock \emph{Oper. Res.}, 60\penalty0 (6):\penalty0 1404–1422, nov 2012.
\newblock ISSN 0030-364X.

\bibitem[Kedia et~al.(2020)Kedia, Jain, and Sharma]{kedia2020price}
Sajan Kedia, Samyak Jain, and Abhishek Sharma.
\newblock Price optimization in fashion e-commerce, 2020.

\bibitem[Streeck et~al.(2024)Streeck, Gellert, Schmitt, Dipkaya, Fux,
  Januschowski, and Berthold]{streeck2024tricks}
Robert Streeck, Torsten Gellert, Andreas Schmitt, Asya Dipkaya, Vladimir Fux,
  Tim Januschowski, and Timo Berthold.
\newblock Tricks from the trade for large-scale markdown pricing: Heuristic cut
  generation for lagrangian decomposition, 2024.

\bibitem[Bertsimas and Kallus(2020)]{Bertsimas2020}
Dimitris Bertsimas and Nathan Kallus.
\newblock From predictive to prescriptive analytics.
\newblock \emph{Manag. Sci.}, 66\penalty0 (3):\penalty0 1025--1044, 2020.
\newblock \doi{10.1287/MNSC.2018.3253}.

\bibitem[Elmachtoub and Grigas(2022)]{Elmachtoub2022}
Adam~N. Elmachtoub and Paul Grigas.
\newblock Smart "predict, then optimize".
\newblock \emph{Manag. Sci.}, 68\penalty0 (1):\penalty0 9--26, 2022.
\newblock \doi{10.1287/MNSC.2020.3922}.

\bibitem[Loh et~al.(2022)Loh, Khandelwal, Regan, and Little]{loh2022promotheus}
Eleanor Loh, Jalaj Khandelwal, Brian Regan, and Duncan~A. Little.
\newblock Promotheus: An end-to-end machine learning framework for optimizing
  markdown in online fashion e-commerce, 2022.

\bibitem[Ferreira et~al.(2016{\natexlab{b}})Ferreira, Lee, and
  Simchi-Levi]{ferreira2016}
Kris~Johnson Ferreira, Bin Hong~Alex Lee, and David Simchi-Levi.
\newblock {Analytics for an Online Retailer: Demand Forecasting and Price
  Optimization}.
\newblock \emph{Manufacturing \& Service Operations Management}, 18\penalty0
  (1):\penalty0 69--88, 2016{\natexlab{b}}.
\newblock \doi{10.1287/msom.2015.0561}.

\bibitem[Collins and Thomas(2012)]{reinforcmentlearningforpricing}
Andrew Collins and Lyn Thomas.
\newblock Comparing reinforcement learning approaches for solving game
  theoretic models: A dynamic airline pricing game example.
\newblock \emph{Journal of the Operational Research Society}, 63:\penalty0
  1165--1173, 08 2012.
\newblock \doi{10.1057/jors.2011.94}.

\bibitem[Petropoulos et~al.(2022)Petropoulos, Apiletti, Assimakopoulos, Babai,
  Barrow, Taieb, Bergmeir, Bessa, Bijak, Boylan,
  et~al.]{petropoulos2022forecasting}
Fotios Petropoulos, Daniele Apiletti, Vassilios Assimakopoulos, Mohamed~Zied
  Babai, Devon~K Barrow, Souhaib~Ben Taieb, Christoph Bergmeir, Ricardo~J
  Bessa, Jakub Bijak, John~E Boylan, et~al.
\newblock Forecasting: theory and practice.
\newblock \emph{International Journal of Forecasting}, 38\penalty0
  (3):\penalty0 705--871, 2022.
\newblock \doi{10.1016/j.ijforecast.2021.11.001}.

\bibitem[Makridakis et~al.(2022)Makridakis, Spiliotis, and
  Assimakopoulos]{M5Results}
Spyros Makridakis, Evangelos Spiliotis, and Vassilios Assimakopoulos.
\newblock M5 accuracy competition: Results, findings, and conclusions.
\newblock \emph{International Journal of Forecasting}, 38\penalty0
  (4):\penalty0 1346--1364, 2022.
\newblock ISSN 0169-2070.
\newblock \doi{10.1016/j.ijforecast.2021.11.013}.

\bibitem[B\"{o}se et~al.(2017)B\"{o}se, Flunkert, Gasthaus, Januschowski,
  Lange, Salinas, Schelter, Seeger, and Wang]{TimDemandForecasting}
Joos-Hendrik B\"{o}se, Valentin Flunkert, Jan Gasthaus, Tim Januschowski,
  Dustin Lange, David Salinas, Sebastian Schelter, Matthias Seeger, and Yuyang
  Wang.
\newblock Probabilistic demand forecasting at scale.
\newblock \emph{Proc. VLDB Endow.}, 10\penalty0 (12):\penalty0 1694–1705,
  2017.
\newblock ISSN 2150-8097.
\newblock \doi{10.14778/3137765.3137775}.
\newblock URL \url{https://doi.org/10.14778/3137765.3137775}.

\bibitem[Tichy et~al.(2024)Tichy, Babounikau, Wolke, Ulbrich, and
  Feindt]{Tichy}
Malte Tichy, Iliau Babounikau, Nikolas Wolke, Stefan Ulbrich, and Michael
  Feindt.
\newblock Scaling-aware rating of count forecasts, 2024.

\bibitem[Friedman(2001)]{friedman2001greedy}
Jerome~H. Friedman.
\newblock Greedy function approximation: A gradient boosting machine.
\newblock \emph{The Annals of Statistics}, 29\penalty0 (5), 2001.
\newblock ISSN 0090-5364.
\newblock \doi{10.1214/aos/1013203451}.

\bibitem[Ke et~al.(2017)Ke, Meng, Finley, Wang, Chen, Ma, Ye, and
  Liu]{ke2017lightgbm}
Guolin Ke, Qi~Meng, Thomas Finley, Taifeng Wang, Wei Chen, Weidong Ma, Qiwei
  Ye, and Tie{-}Yan Liu.
\newblock Lightgbm: {A} highly efficient gradient boosting decision tree.
\newblock In Isabelle Guyon, Ulrike von Luxburg, Samy Bengio, Hanna~M. Wallach,
  Rob Fergus, S.~V.~N. Vishwanathan, and Roman Garnett, editors, \emph{Advances
  in Neural Information Processing Systems 30: Annual Conference on Neural
  Information Processing Systems 2017, December 4-9, 2017, Long Beach, CA,
  {USA}}, pages 3146--3154, 2017.

\bibitem[Lainder and Wolfinger(2022)]{M5UncertaintyResults}
A.~David Lainder and Russell~D. Wolfinger.
\newblock Forecasting with gradient boosted trees: augmentation, tuning, and
  cross-validation strategies: Winning solution to the m5 uncertainty
  competition.
\newblock \emph{International Journal of Forecasting}, 38\penalty0
  (4):\penalty0 1426--1433, 2022.
\newblock ISSN 0169-2070.
\newblock \doi{https://doi.org/10.1016/j.ijforecast.2021.12.003}.
\newblock URL
  \url{https://www.sciencedirect.com/science/article/pii/S0169207021002090}.
\newblock Special Issue: M5 competition.

\bibitem[Schultz et~al.(2024)Schultz, Stephan, Sieber, Yeh, Kunz, Doupe, and
  Januschowski]{schultz2024causal}
Douglas Schultz, Johannes Stephan, Julian Sieber, Trudie Yeh, Manuel Kunz,
  Patrick Doupe, and Tim Januschowski.
\newblock Causal forecasting for pricing, 2024.

\bibitem[Birr and Gouttes(2023)]{prato_talk2023}
Stefan Birr and Adele Gouttes.
\newblock Hierarchical forecasts: A case study from pricing in e-commerce.
\newblock 2023 IIF Workshop on Forecast Reconciliation, 2023.
\newblock URL
  \url{https://robjhyndman.com/files/prato/Hierarchical%20Forecasting%20Review.pdf}.

\bibitem[Sprangers et~al.(2024)Sprangers, Wadman, Schelter, and
  de~Rijke]{Sprangers2024}
Olivier Sprangers, Wander Wadman, Sebastian Schelter, and Maarten de~Rijke.
\newblock Hierarchical forecasting at scale.
\newblock \emph{International Journal of Forecasting}, 2024.
\newblock \doi{10.1016/j.ijforecast.2024.02.006}.

\bibitem[Petropoulos et~al.(2025)Petropoulos, Hollyman, and
  Spiliotis]{Petropoulos12122025}
Fotios Petropoulos, Ross~A Hollyman, and Evangelos Spiliotis.
\newblock Scalable forecast reconciliation through (un)guided sub-hierarchies.
\newblock \emph{Journal of the Operational Research Society}, 0\penalty0
  (0):\penalty0 1--20, 2025.
\newblock \doi{10.1080/01605682.2025.2599394}.
\newblock URL \url{https://doi.org/10.1080/01605682.2025.2599394}.

\bibitem[Huelden et~al.(2024)Huelden, Jascisens, Roemheld, and
  Werner]{Huelden2024}
Tobias Huelden, Vitalijs Jascisens, Lars Roemheld, and Tobias Werner.
\newblock Human-machine interactions in pricing: Evidence from two large-scale
  field experiments.
\newblock \emph{SSRN Electronic Journal}, 2024.
\newblock ISSN 1556-5068.
\newblock \doi{10.2139/ssrn.4763132}.

\bibitem[Ehrgott(2005)]{Ehrgott2005}
Matthias Ehrgott.
\newblock \emph{Multicriteria Optimization}.
\newblock Springer, 2005.
\newblock ISBN 978-3-540-21398-7.
\newblock \doi{10.1007/3-540-27659-9}.

\bibitem[Marler and Arora(2010)]{marler2010weighted}
R~Timothy Marler and Jasbir~S Arora.
\newblock The weighted sum method for multi-objective optimization: new
  insights.
\newblock \emph{Structural and multidisciplinary optimization}, 41:\penalty0
  853--862, 2010.

\bibitem[Chen et~al.(2023)Chen, Li, Yoder, Arik, and
  Pfister]{chen2023tsmixerallmlparchitecturetime}
Si-An Chen, Chun-Liang Li, Nate Yoder, Sercan~O. Arik, and Tomas Pfister.
\newblock Tsmixer: An all-mlp architecture for time series forecasting, 2023.
\newblock URL \url{https://arxiv.org/abs/2303.06053}.

\bibitem[Januschowski et~al.(2022)Januschowski, Wang, Torkkola, Erkkilä,
  Hasson, and Gasthaus]{Januschowski2022}
Tim Januschowski, Yuyang Wang, Kari Torkkola, Timo Erkkilä, Hilaf Hasson, and
  Jan Gasthaus.
\newblock Forecasting with trees.
\newblock \emph{International Journal of Forecasting}, 38\penalty0
  (4):\penalty0 1473--1481, 2022.
\newblock ISSN 0169-2070.
\newblock \doi{10.1016/j.ijforecast.2021.10.004}.

\bibitem[Abadie(2005)]{abadie2005}
Alberto Abadie.
\newblock Semiparametric difference-in-differences estimators.
\newblock \emph{The review of economic studies}, 72\penalty0 (1):\penalty0
  1--19, 2005.

\end{thebibliography}
